\documentclass{article}

\usepackage{amsmath}
\usepackage{graphicx}
\usepackage{caption}
\usepackage{float}
\usepackage{enumitem}
\usepackage{pgfplots}
\pgfplotsset{compat=1.16}
\usepgfplotslibrary{groupplots}

\usepackage[preprint]{corl_2026} 

\newcommand{\gain}[1]{\textcolor[rgb]{0,0.55,0}{(+#1)}}
\newcommand{\loss}[1]{\textcolor[rgb]{0.80,0.10,0.10}{($-$#1)}}

\definecolor{cOurs}{RGB}{99,137,206}
\definecolor{cAux}{RGB}{212,147,148}
\definecolor{cVanilla}{RGB}{165,160,152}

\title{Gaze Prompts: Temporally Dense Human Attention for Vision--Language--Action Fine-Tuning}

\author{
  Yihan Zhou$^{1,2,3 *}$, 
  Rui Yan$^{1 *}$, 
  Mingcong Li$^{4}$, 
  Zheyuan Huang$^{4}$, \\
  Xu Yang$^{1}$, 
  Xueyang Guo$^{1,2,3}$, 
  Yilin Mo$^{1,2,3 \star}$ \\[0.6em]
  $^{1}$Department of Automation, Tsinghua University \\
  $^{2}$Beijing Key Laboratory of Embodied Intelligence Systems \\
  $^{3}$Institute for Embodied Intelligence and Robotics, Tsinghua University \\
  $^{4}$LingYu Robotics \\[0.4em]
  $^{*}$Equal Contribution \qquad
  $^{\star}$Corresponding Author \\[0.4em]
  \texttt{\{zhouyh23, yan-r22\}@mails.tsinghua.edu.cn}, \
  \texttt{ylmo@tsinghua.edu.cn}
}

\hypersetup{
  pdftitle={Gaze Prompts: Temporally Dense Human Attention for Vision-Language-Action Fine-Tuning},
  pdfauthor={Yihan Zhou, Rui Yan, Mingcong Li, Zheyuan Huang, Xu Yang, Xueyang Guo, Yilin Mo}
}

\begin{document}
\maketitle


\begin{abstract}
Vision--Language--Action (VLA) fine-tuning pairs images with actions at every step, yet typically provides only a task-level language instruction, leaving moment-to-moment visual relevance implicit. We introduce \emph{eye-tracker-supervised gaze prompting}, which uses gaze recorded during VR teleoperation to provide frame-level visual guidance for VLA fine-tuning. During training, recorded gaze locations are rendered as crosshairs on the robot's head-camera images. At deployment, a lightweight predictor estimates gaze locations from recent images and the instruction, supplying the same type of visual prompt without an eye tracker or changes to the policy architecture. Instantiated with $\pi_0$, gaze prompting increases mean success from $26.3\%$ to $56.0\%$ across six real-world bimanual manipulation tasks, with gains also observed when a single policy is trained on all six tasks. We release \textsc{GazeMani}, a dataset of $1{,}200$ teleoperated trajectories with synchronized gaze.
\end{abstract}




\begin{figure}[h]
    \centering
    \includegraphics[width=\linewidth]{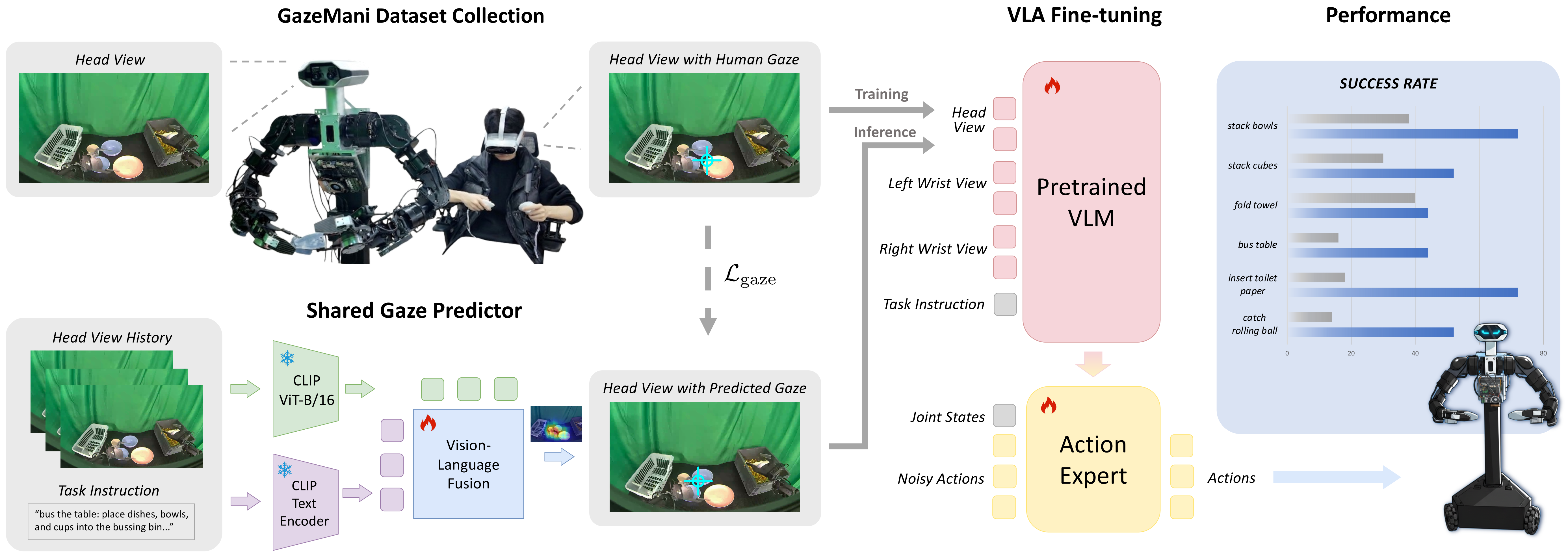}
    \caption{\textbf{Eye-tracker-supervised gaze prompting.} During VR teleoperation (upper left), an HMD eye tracker records the operator's gaze, which supervises a lightweight gaze predictor (lower left) and is rendered as a crosshair on the head-camera image for VLA fine-tuning (center). At deployment, the predictor estimates gaze from recent head-camera frames and the language instruction, providing the same visual prompt to the fine-tuned VLA without an eye tracker or architectural modifications. The right panel summarizes task performance.}
    \label{fig:overview}
\end{figure}

\section{Introduction}
\label{sec:intro}

Vision--Language--Action (VLA) models couple pretrained vision--language backbones with large-scale robot pretraining~\citep{zitkovich2023rt,kim2024openvla,black2024pi_0,intelligence2025pi_}, and are then adapted to a new task or platform by fine-tuning on a limited number of demonstrations. The supervision available during fine-tuning, however, arrives at two very different granularities: each demonstration pairs images with an action at every frame, while its language instruction specifies the goal only once. Successful execution requires frame-by-frame choices about which object instance to attend, which task phase is active, and where contact will occur next, yet neither signal explicitly annotates these frame-level decisions. What is missing is a temporally dense trace of what the demonstrator considered relevant at each moment.

The consequences of this gap are increasingly well documented. Mechanistic analyses show that language does not uniformly drive VLA action generation: actions depend predominantly on visual and spatial features, with instruction changes affecting behavior mainly when the scene alone underdetermines the correct action~\citep{grant2026not}. Fine-tuning can reinforce this imbalance, since robot data offers dense visual--action supervision but little linguistic variation, encouraging shortcut learning and language forgetting~\citep{xu2025seeing}. Existing remedies inject spatial structure through visual prompts, auxiliary losses, or attention modifications---but their cues are engineered or model-derived rather than measured from the demonstrator's own attention.

The missing trace is naturally available in VR teleoperation: the operator's headset records their gaze alongside each demonstration through its integrated eye tracker. Prior studies of natural behavior have shown that gaze is closely coupled to action, with fixations selecting task-relevant objects, shifting across sub-task boundaries, and anticipating contact before the hand arrives~\citep{johansson2001eye,land2001ways}. Yet this signal is available only during data collection, since the deployed robot carries no eye tracker. Using gaze at deployment therefore requires predicting it from the robot's own observations.

We introduce \emph{eye-tracker-supervised gaze prompting} for VLA fine-tuning (Figure~\ref{fig:overview}). During fine-tuning, the recorded gaze location is rendered as a crosshair on the VLA's RGB input. A lightweight gaze predictor (${\sim}2.9$M trainable parameters) is trained independently to map RGB observations and language to the gaze location, providing the gaze signal without an eye tracker at deployment. This converts gaze recorded during demonstration collection into a deployable visual cue for the VLA's existing image pathway.

Overall, our contributions are summarized as follows:
\begin{enumerate}[topsep=2pt,itemsep=2pt,leftmargin=*]
\item We use human gaze recorded during teleoperation as temporally dense supervision for VLA fine-tuning, and learn a lightweight predictor that provides gaze prompts to an unchanged policy without an eye tracker or architectural modification at deployment.

\item Experiments on six real-world bimanual manipulation tasks validate the effectiveness of gaze prompting for VLA fine-tuning, with mean success increasing from $26.3\%$ to $56.0\%$ and comparable gains under unified multi-task training.

\item Controlled route and cue-source analyses characterize the contribution of inference-time gaze cues and show that human gaze can indicate upcoming task-relevant locations.

\end{enumerate}

We release the \textsc{GazeMani} dataset, comprising 1{,}200 bimanual trajectories with synchronized gaze recordings across six tasks, along with evaluation code at \url{https://gazemani.github.io/}.



\section{Related Work}
\label{sec:related_work}

\paragraph{Language conditioning in VLA fine-tuning.}
VLA policies inherit broad semantic knowledge from pretrained vision--language models, yet recent studies suggest that language does not always play a central role in action generation. Mechanistic analyses show that actions can rely heavily on visual and spatial features, while language becomes more influential when the scene alone leaves multiple plausible actions~\citep{grant2026not}. Fine-tuning can further amplify this imbalance: robot datasets provide dense visual--action supervision but comparatively little linguistic variation, encouraging shortcut learning, information collapse, and language forgetting~\citep{xu2025seeing,lian2026langforce}. Other work addresses this mismatch through guidance, decomposition, or alignment objectives~\citep{zhan2026stable,wulff2026grounding,huang2026thinkact}. Our work takes a complementary approach by augmenting VLA fine-tuning with a temporally dense signal of the demonstrator's visual attention, rather than modifying the language objective or adding a reasoning module.

\paragraph{Spatial and attentional cues for VLA policies.}
Existing methods inject task-relevant visual structure into VLA policies through several routes. Auxiliary-loss methods regularize internal attention toward gaze, saliency, or grounded regions~\citep{pani2026gaze,song2026reconvla}; architectural methods modify how attention is computed~\citep{zhang2026focusvla,xiao2025ava}; and inference-time guidance methods use modified visual inputs to stabilize object-centric action generation~\citep{zhou2026tag}. A closely related line renders spatial cues directly onto the RGB observation. TraceVLA draws past motion traces~\citep{zheng2025tracevla}; AimBot renders scope reticles encoding end-effector pose~\citep{dai2025aimbot}; HAMSTER and RT-Affordance provide VLM- or affordance-derived paths and waypoints~\citep{li2025hamster,nasiriany2025rt}; and VP-VLA renders planner-generated crosshairs and boxes~\citep{wang2026vp}. Across these approaches, the attentional or spatial guidance is typically engineered or model-derived, drawing on robot geometry, scene heuristics, external planners, model internals, or synthetic gaze rather than the human demonstrator's gaze.

\paragraph{Human gaze as an attentional signal.}
Studies of natural behavior show that gaze is tightly coupled to action: fixations select task-relevant objects, shift across sub-task boundaries, and anticipate upcoming contact~\citep{land2001ways,hayhoe2005eye,land2000eye}. This has motivated imitation-learning methods that use gaze through feature masks, gaze-centered crops, auxiliary losses, or additional input channels~\citep{zhang2018agil,kim2020using,saran2020efficiently,thammineni2023selective}, as well as more recent robot and VLA methods for foveated tokenization, attention regularization, point-cloud cropping, learned eye policies, and subtask decomposition~\citep{chuang2025look,pani2026gaze,kerr2025eye,takizawa2025gaze}. Our work connects prior work on human gaze with pixel-space visual prompting: gaze recorded during teleoperation provides supervision for a lightweight predictor, whose predicted gaze location is delivered as a per-frame RGB prompt through the existing visual pathway of an unchanged VLA policy.



\section{Method}
\label{sec:method}

Our method turns gaze recorded during teleoperated demonstrations into a deployable pixel-space prompt for a VLA policy. During policy fine-tuning, the recorded gaze location is rendered as a crosshair on the head-camera image. Separately, a lightweight predictor learns to estimate gaze from recent head-camera frames and the language instruction; at inference, its predicted gaze location is rendered in the same way, providing a per-frame gaze cue without an eye tracker. The policy architecture is unchanged, and the gaze cue reaches action generation only through the same visual pathway that processes the rest of the image.

The pipeline has three components. A gaze predictor (${\sim}$2.9M trainable parameters) maps a short window of $N$ head-camera frames and the language instruction to a gaze heatmap over the current view (\S\ref{sec:method:predictor}). A renderer converts a gaze location into a crosshair on the head-camera image, while wrist-camera views remain unchanged (\S\ref{sec:method:rendering}). A VLA policy, instantiated as $\pi_0$~\citep{black2024pi_0} in our experiments, is fine-tuned on crosshair-rendered observations, consumes the resulting multi-view input, and emits actions. The predictor and policy are trained separately (\S\ref{sec:method:training_inference}).

\subsection{Gaze predictor}
\label{sec:method:predictor}

\begin{figure}[t]
    \centering
    \includegraphics[width=0.8\linewidth]{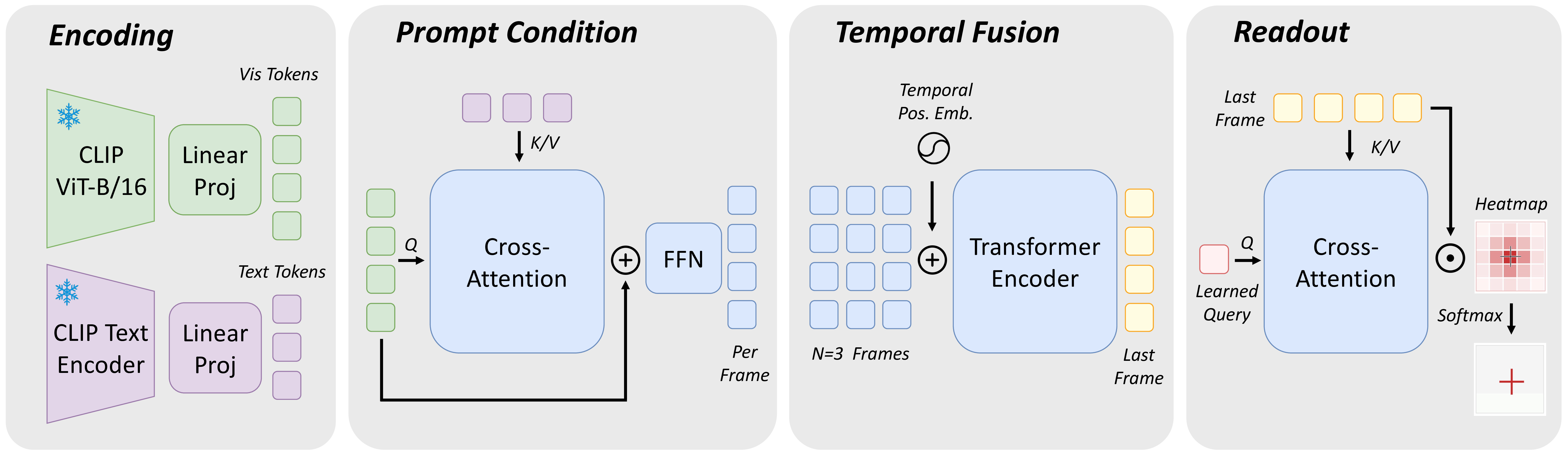}
    \caption{Gaze predictor architecture. Frozen CLIP encoders encode each head-camera frame and the instruction, and learned linear projections map their tokens to a shared fusion space (\textbf{Encoding}); each frame's visual tokens cross-attend to the text tokens (\textbf{Prompt conditioning}); a transformer fuses the $N{=}3$ prompt-conditioned frames and retains the last frame's tokens (\textbf{Temporal fusion}); these are reshaped to a $14{\times}14$ grid and upsampled to a $224{\times}224$ gaze heatmap (\textbf{Readout}). The architecture details are given in \S\ref{sec:method:predictor}.}
    \label{fig:predictor_arch}
    \vspace{-10pt}
\end{figure}

The predictor maps $N$ consecutive head-camera frames $\mathbf{I}_{t\text{-}N{+}1:t}$ and a language instruction $\ell$ to a $224\times 224$ gaze heatmap over the current view (Figure~\ref{fig:predictor_arch}); we use $N=3$ throughout.

\paragraph{Architecture.}
A frozen CLIP ViT-B/16~\citep{radford2021learning} vision encoder processes each frame independently into $196$ visual tokens (one per $16{\times}16$ patch); a frozen CLIP text encoder embeds the instruction into $T$ text tokens. Both token sequences are linearly projected to a shared 256-dimensional fusion space. We keep the two CLIP encoders frozen and train everything else---the projections, the prompt-conditioning and temporal-fusion blocks, and the readout decoder. After encoding, the architecture has three trainable stages:
\begin{enumerate}[topsep=2pt,itemsep=2pt,leftmargin=*]
    \item \emph{Per-frame prompt conditioning.} Each frame's visual tokens cross-attend to the text tokens (vision queries text), followed by a feed-forward network. This injects instruction semantics into every frame's spatial representation before any temporal mixing.
    \item \emph{Temporal fusion.} The $N$ prompt-conditioned token grids ($N{\times}196$ tokens) are augmented with learnable temporal positional embeddings and processed by a 2-layer transformer encoder. We extract the last frame's contextualized tokens as the output; these now carry information from the preceding $N{-}1$ frames.
    \item \emph{Readout.} The last frame's $196$ contextualized tokens are reshaped to a $14\times14$ feature map and decoded by a learned transposed-convolution upsampler into $224\times224$ logits, one score per output cell. Decoding to $224\times224$ provides a denser spatial readout than predicting gaze coordinates directly on the coarse $14\times14$ grid.
\end{enumerate}

\paragraph{Supervision.}
The ground-truth gaze location is the operator's eye-tracker gaze registered to the head-camera view and temporally filtered during dataset preprocessing. Each frame's gaze location is rasterized into a normalized Gaussian-smoothed $224\times 224$ heatmap $G^{\star}$ (smoothing $\sigma$ ablated in Appendix~\ref{app:sigma}). The predictor is supervised with per-frame cross-entropy:
\begin{equation}
    \mathcal{L}_{\text{gaze}} \;=\; -\sum_{i,j} G^{\star}_{ij}\,\log \hat{G}_{ij}, \qquad
    \hat{G} \;=\; \mathrm{softmax}\!\left(f_\theta(\mathbf{I}_{t\text{-}N{+}1:t},\, \ell)\right),
    \label{eq:gaze_loss}
\end{equation}
where $f_\theta$ denotes the full predictor producing unnormalized $224\times224$ logits. The predictor is trained jointly on demonstrations from all six tasks, exposing it to a broader range of task and visual contexts than any single task. Because per-task episode lengths differ substantially, we use a task-balanced weighted sampler so that each of the six tasks receives equal sampling probability and longer tasks do not dominate training.

\subsection{Visual-prompt rendering}
\label{sec:method:rendering}

The renderer draws a crosshair at a given gaze location on the head-camera image, with fixed geometry and color across all tasks and instructions (details in Appendix~\ref{app:crosshair}). Only the head view is modified; wrist views pass through unchanged. During policy fine-tuning, the crosshair is rendered at the ground-truth gaze. At inference, the gaze location is obtained from the predictor's $224\times224$ heatmap by applying a soft-argmax over the $3\times3$ neighborhood of its peak, followed by causal exponential moving-average (EMA) smoothing of the resulting coordinates.

\subsection{Training and inference}
\label{sec:method:training_inference}

The predictor and policy are trained independently, with no gradients flowing between them. The predictor is supervised by ground-truth gaze locations via Eq.~\eqref{eq:gaze_loss}. The policy is fine-tuned with the crosshair rendered at the ground-truth gaze location, while the action target remains the original demonstrated action. At inference, the frozen predictor estimates gaze from a rolling buffer of the $N$ most recent head-camera frames, and the VLA policy acts on the resulting crosshair-rendered observation. The predictor adds one forward pass per policy step (latency in Appendix~\ref{app:latency}).



\section{Experimental Results}
\label{sec:result}

\subsection{Setup}
\label{sec:result:setup}

\begin{figure}[t]
    \centering
    \includegraphics[width=\linewidth]{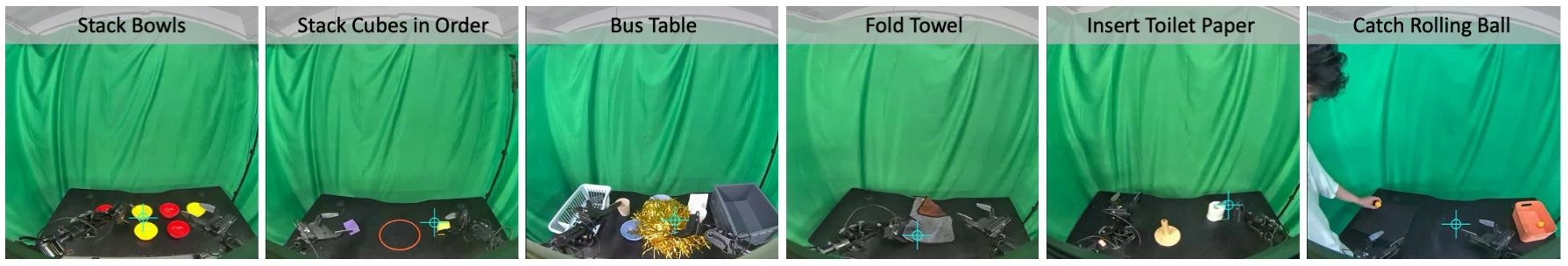}
    \caption{\textbf{The \textsc{GazeMani} task suite.} Six real-world bimanual manipulation tasks used for policy evaluation; the cyan crosshair is the rendered gaze prompt.}
    \label{fig:task_suite}
    \vspace{-13pt}
\end{figure}

\paragraph{Task suite.}
We evaluate on six real-world bimanual manipulation tasks (Figure~\ref{fig:task_suite}; full descriptions in Appendix~\ref{app:tasks}). The suite spans instruction-conditioned instance selection (\emph{Stack Bowls}, red or yellow), phase-dependent sequencing (\emph{Stack Cubes in Order}), long-horizon tidying (\emph{Bus Table}), deformable-state tracking (\emph{Fold Towel}), precise contact localization (\emph{Insert Toilet Paper}), and dynamic interception (\emph{Catch Rolling Ball}).

\paragraph{Robot platform.}
We use TeleAvatar, a bimanual mobile manipulator from LingYu Robotics, equipped with two 7-DoF arms, one-DoF grippers, a head-mounted stereo fisheye camera, and two wrist cameras. The platform is used for both teleoperated data collection and policy deployment (Figure~\ref{fig:platform}); additional hardware details are provided in Appendix~\ref{app:platform}.

\begin{figure}[h]
    \centering
    \includegraphics[width=0.37\linewidth]{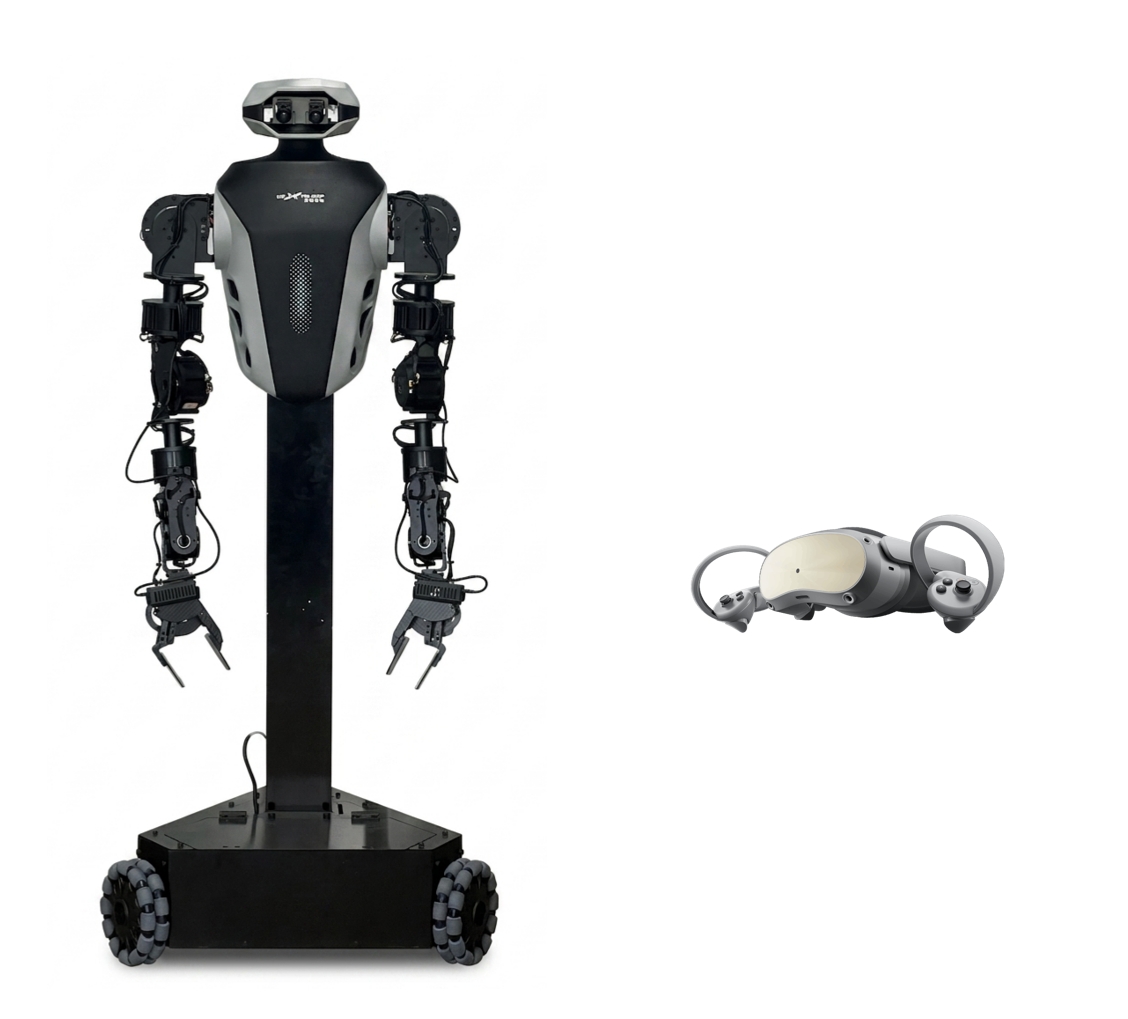}
    \caption{Experimental platform, used for both teleoperated data collection and policy deployment. \emph{Left:} the bimanual mobile manipulator---two 7-DoF arms with one-DoF grippers on a torso column above a wheeled base, with a head-mounted stereo fisheye camera. \emph{Right:} the Pico~4 Enterprise headset and its two controllers, used only during data collection for VR teleoperation and eye tracking---no headset or eye tracker is needed at deployment.}
    \label{fig:platform}
\end{figure}

\paragraph{Data collection (\textsc{GazeMani}).}
\label{sec:result:data}
Demonstrations are collected via VR teleoperation using a Pico 4 Enterprise headset with an integrated eye tracker. The recorded 3D gaze rays are registered offline to 2D head-camera image coordinates (Appendix~\ref{app:gaze_projection}), and the resulting gaze trace is temporally smoothed with a Kalman filter. Each frame records three RGB views, the robot's proprioceptive joint state, the demonstrated action, and the operator's gaze location. Each task provides 200 demonstrations (1{,}200 total); per-task counts and episode-length statistics are reported in Appendix~\ref{app:dataset_stats}.

\paragraph{Methods.}
We compare three methods that share the same $\pi_0$ backbone and demonstration set (training and hyperparameter details in Appendix~\ref{app:training}):
\begin{itemize}[topsep=2pt,itemsep=2pt,leftmargin=*]
    \item \textbf{Vanilla $\pi_0$}: standard fine-tuning on unmodified RGB observations; no gaze signal.
    \item \textbf{Aux-loss $\pi_0$}: the Gaze-Regularized VLA mechanism~\citep{pani2026gaze}, which adds an auxiliary KL loss regularizing the policy's language$\to$vision attention toward the gaze location while leaving the architecture and action objective unchanged. We replace the original synthetic gaze with our recorded human gaze and keep the original KL coefficient. The gaze signal is available only during training.
    \item \textbf{Gaze-prompt $\pi_0$ (ours)}: $\pi_0$ fine-tuned on head-view images with a crosshair rendered at the recorded gaze location, and evaluated with a crosshair rendered at the predicted gaze location. The policy architecture is unchanged.
\end{itemize}
We fine-tune a separate policy for each task in the main experiments and additionally evaluate a single policy trained jointly across all six tasks in \S\ref{sec:result}.
Object placements are randomized across rollouts, with fixed seeds for stochastic policy components; rollout counts are reported within each experiment.

\subsection{Policy performance}
\label{sec:result:gain}

Table~\ref{tab:main_results} reports the main results with a separately fine-tuned policy for each task. Gaze-prompt $\pi_0$ improves over vanilla on all six tasks, increasing mean success from $26.3\%$ to $56.0\%$ ($+29.7$ points), and outperforms aux-loss on every task. The largest gains over vanilla occur on T1, T5, and T6, where successful execution requires selecting or precisely localizing small, visually ambiguous, or moving targets; the gain is smallest on Fold Towel (T4), where vanilla already reaches $42\%$. Two-sided Fisher exact tests find the improvement over vanilla significant on five of the six tasks, with T4 as the exception. To account for differences in baseline performance, the last column additionally reports the headroom-normalized gain $\Delta_{\text{norm}}=\Delta_{\text{SR}}/(1-\text{SR}_{\text{vanilla}})$. Uncut single-take recordings of all rollouts reported in Table~\ref{tab:main_results} are available on the project website.

\begin{table}[t]
    \centering
    \small
    \caption{Main policy results across six manipulation tasks. Cells report success rate over 50 rollouts (best per task in bold). Parenthesized values are the change over vanilla in percentage points (\textcolor[rgb]{0,0.55,0}{green}: gain, \textcolor[rgb]{0.80,0.10,0.10}{red}: loss); the last column is the headroom-normalized gain $\Delta_{\text{norm}} = \Delta\text{SR}/(1-\text{SR}_{\text{vanilla}})$ of gaze-prompt over vanilla.}
    \label{tab:main_results}
    \begin{tabular}{llcccc}
        \hline
        Task & Description & Vanilla & Aux-loss & Gaze-prompt & $\Delta_{\text{norm}}$ \\
        \hline
        T1 & Stack Bowls          & 38\% & 60\% \gain{22} & \textbf{72\%} \gain{34} & 55\% \\
        T2 & Stack Cubes in Order & 30\% & 44\% \gain{14} & \textbf{52\%} \gain{22} & 31\% \\
        T3 & Bus Table            & 16\% &  8\% \loss{8}  & \textbf{44\%} \gain{28} & 33\% \\
        T4 & Fold Towel           & 42\% & 20\% \loss{22} & \textbf{44\%} \gain{2}  &  3\% \\
        T5 & Insert Toilet Paper  & 18\% & 44\% \gain{26} & \textbf{72\%} \gain{54} & 66\% \\
        T6 & Catch Rolling Ball   & 14\% & 24\% \gain{10} & \textbf{52\%} \gain{38} & 44\% \\
        \hline
        \multicolumn{2}{l}{Average} & 26.3\% & 33.3\% \gain{7.0} & \textbf{56.0\%} \gain{29.7} & --- \\
        \hline
    \end{tabular}
\end{table}

\begin{figure}[t]
    \centering
    \scalebox{0.9}{%
    \begin{tikzpicture}
    \begin{groupplot}[
        group style={group size=2 by 1, horizontal sep=1.5cm},
        width=0.46\linewidth, height=5cm,
        ymin=0, ymax=100,
        ylabel={Success rate (\%)},
        ymajorgrids, tick label style={font=\scriptsize},
        label style={font=\small}, title style={font=\small},
    ]
    \nextgroupplot[
        title={(a) Fold Towel (sequential)},
        xtick={1,2,3,4,5,6},
        xticklabels={1st grasp,2nd grasp,1st fold,rotate,2nd fold,3rd fold},
        x tick label style={rotate=35,anchor=east,font=\scriptsize},
        legend style={font=\scriptsize,at={(0.03,0.04)},anchor=south west,draw=none,fill=none},
    ]
    \addplot[mark=*,thick,cVanilla] coordinates {(1,90)(2,70)(3,62)(4,62)(5,52)(6,42)};
    \addlegendentry{Vanilla}
    \addplot[mark=triangle*,thick,cAux] coordinates {(1,90)(2,78)(3,58)(4,56)(5,26)(6,20)};
    \addlegendentry{Aux-loss}
    \addplot[mark=square*,thick,cOurs] coordinates {(1,90)(2,70)(3,50)(4,48)(5,44)(6,44)};
    \addlegendentry{Gaze-prompt (ours)}
    \nextgroupplot[
        title={(b) Bus Table (per item)},
        ybar, bar width=3pt, enlarge x limits=0.1,
        symbolic x coords={tinsel,pcup,pbag,plate,lgbowl,smbowl,cup},
        xtick=data,
        xticklabels={pom-pom,paper cup,paper bag,plate,large bowl,small bowl,cup},
        x tick label style={rotate=35,anchor=east,font=\scriptsize},
    ]
    \addplot[fill=cVanilla,draw=cVanilla] coordinates {(tinsel,96)(pcup,60)(pbag,60)(plate,42)(lgbowl,32)(smbowl,30)(cup,18)};
    \addplot[fill=cAux,draw=cAux] coordinates {(tinsel,98)(pcup,66)(pbag,70)(plate,42)(lgbowl,30)(smbowl,22)(cup,10)};
    \addplot[fill=cOurs,draw=cOurs] coordinates {(tinsel,98)(pcup,84)(pbag,72)(plate,64)(lgbowl,52)(smbowl,56)(cup,50)};
    \end{groupplot}
    \end{tikzpicture}}
    \caption{\textbf{Stage-wise success on the long-horizon tasks.}
    (a) Fold Towel, per fold stage; (b) Bus Table, per item.}
    \label{fig:stagewise}
\end{figure}
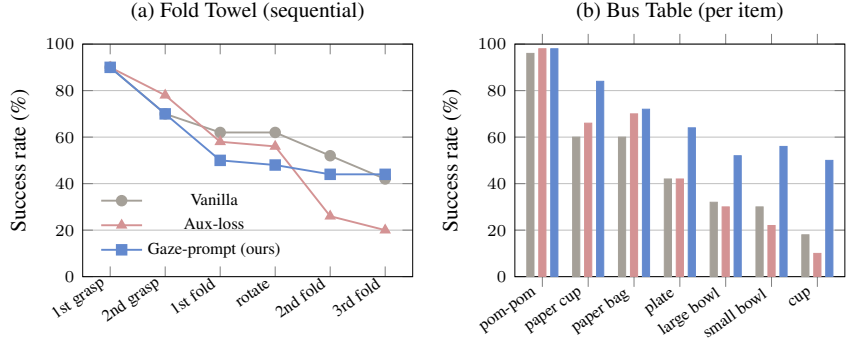

The two longest tasks, T3 and T4, further show how performance evolves within a rollout (Figure~\ref{fig:stagewise}). On Fold Towel, gaze-prompt remains close to vanilla across the sequence, consistent with their similar final success rates. On Bus Table, the advantage of gaze-prompt is more pronounced on the later, harder items, sorting $4.8$ of $7$ items on average compared with $3.4$ for both baselines.

\begin{figure}[t]
    \centering
    \includegraphics[width=0.5\linewidth]{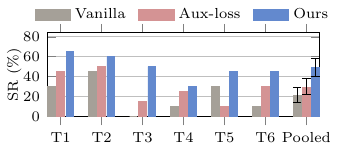}
    \caption{\textbf{Unified multi-task policy performance.} Each method is trained as a single policy jointly across all six tasks. Bars show observed per-task success rates over 20 rollouts. Pooled results aggregate 120 rollouts per method and include 95\% Wilson confidence intervals.}
    \label{fig:multitask}
\end{figure}

The gain also persists when all six tasks are learned by a single policy (Figure~\ref{fig:multitask}). Gaze-prompt achieves the highest observed success rate on all six tasks and reaches $49.2\%$ pooled success over 120 rollouts, compared with $20.8\%$ for vanilla and $29.2\%$ for aux-loss. Under unified training, its improvement over vanilla remains comparable to that under per-task training. These results show that the benefit of gaze prompting is not specific to training an independent policy for each task.

\subsection{Injection route and inference-time cue}
\label{sec:result:attention}
The two gaze-supervised variants incorporate human gaze in different ways. Aux-loss uses gaze through a training-time attention objective, whereas gaze-prompt renders the gaze location directly in the visual input and continues to provide the predicted cue at inference. To examine the contribution of this inference-time cue, we evaluate the same T1 gaze-prompt policy with the crosshair removed only at inference. Success decreases from $72\%$ to $58\%$ over 50 rollouts, close to the $60\%$ achieved by aux-loss. The observed drop is consistent with a contribution from the fresh per-frame cue available at inference.

An instruction-swap analysis on Stack Bowls (T1) further shows how this cue affects the policy's response to language. The visual scene is kept unchanged while the instruction switches between stacking the red and yellow bowls. Under this controlled swap, vanilla $\pi_0$ distributes attention broadly across the tabletop, whereas the gaze-supervised variants show stronger instruction-dependent shifts in attention (Figure~\ref{fig:attn_grid}; extraction details in Appendix~\ref{app:attention_extraction}). Gaze-prompt exhibits the largest instruction-conditioned shift, with a Jensen--Shannon (JS) divergence of $0.117$ between the two attention maps, compared with $0.068$ for aux-loss and $0.047$ for vanilla. Corresponding task success is reported in Table~\ref{tab:swap_prompt}.

\begin{figure*}[t]
    \centering
    \begin{minipage}[c]{0.45\textwidth}
        \centering
        \includegraphics[width=\linewidth]{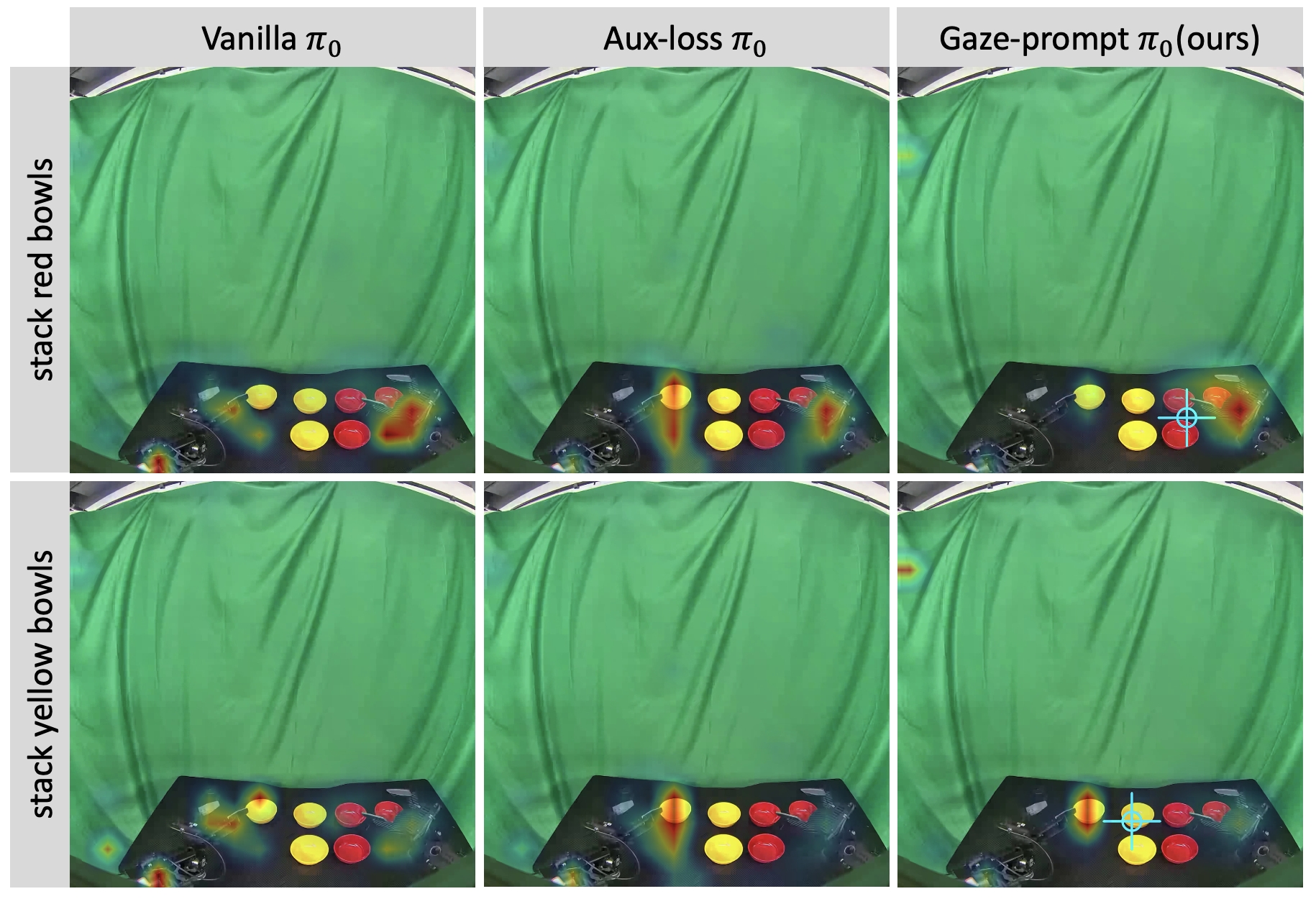}
        \caption{T1 instruction-swap attention.}
        \label{fig:attn_grid}
    \end{minipage}\hfill
    \begin{minipage}[c]{0.5\textwidth}
        \centering
        \captionof{table}{T1 instruction-swap results.}
        \label{tab:swap_prompt}
        \small
        \begin{tabular}{lccc}
            \hline
            Method & SR$_A$ & SR$_B$ & JS \\
            \hline
            Vanilla $\pi_0$ & 40\% & 36\% & 0.047 \\
            Aux-loss $\pi_0$ & 60\% & 60\% & 0.068 \\
            Gaze-prompt $\pi_0$ & \textbf{84\%} & \textbf{60\%} & \textbf{0.117} \\
            \hline
        \end{tabular}

        \vspace{2pt}
        {\footnotesize A: stack red bowls;\quad B: stack yellow bowls}
    \end{minipage}
    \vspace{-13pt}
\end{figure*}

\subsection{Cue source and anticipation}
\label{sec:result:ablation}

To examine how the source of the visual cue affects policy performance, we compare five variants on T6 (\emph{Catch Rolling Ball}) under the same input-pixel rendering route. Each prompted variant independently fine-tunes $\pi_0$ on its corresponding cue. T6 provides a clean setting for this comparison because it contains a single moving target, avoiding an additional object-selection rule when constructing engineered cues. We compare predicted human gaze, a Kalman-extrapolated future-position cue, current-position detection, a random crosshair, and no cue. For the two position-based cues, we use the same CPU color-blob detector on the policy's $224\times224$ input. It runs in $4.06$\,ms, close to the gaze predictor's $3.8$\,ms, and places the cue on the ball in $94.5\%$ $[90.4,96.9]$ of 200 manually checked pre-grasp frames. The Kalman cue extrapolates the detected trajectory $541$\,ms ahead, matching the mean interval from arm-movement onset to the ball's arrival at the movement-onset gaze location.

\begin{table}[t]
    \centering
    \small
    \caption{Cue-source ablation on T6. All prompted variants use the same input-pixel rendering route. Brackets report Wilson 95\% confidence intervals over 50 rollouts.}
    \label{tab:cue_ablation}
    \begin{tabular}{lc}
        \hline
        Cue source & Success rate \\
        \hline
        Human-gaze predictor (ours) & \textbf{52\%} $[38.5,65.2]$ \\
        Kalman future-position cue   & 38\% $[25.9,51.8]$ \\
        Current-position detection  & 12\% $[5.6,23.8]$ \\
        Random crosshair             & 10\% $[4.3,21.4]$ \\
        No cue                       & 14\% $[7.0,26.2]$ \\
        \hline
    \end{tabular}
\end{table}

\begin{figure}[t]
    \centering
    \includegraphics[width=0.98\linewidth]{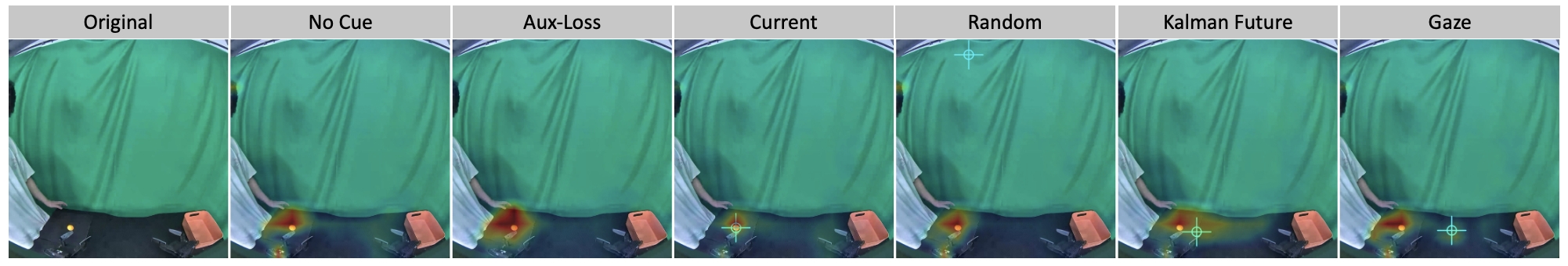}
    \caption{\textbf{Representative attention maps on T6.}
    Columns show the original frame and attention maps for No Cue, Aux-Loss, Current, Random, Kalman Future, and Gaze. Aux-Loss uses human gaze only during training and is shown for reference; the remaining prompted variants use rendered inference-time cues. ``Kalman Future'' extrapolates the detected ball trajectory $541$\,ms ahead. Cyan crosshairs denote rendered cues.}
    \label{fig:t6_attention}
    \vspace{-10pt}
\end{figure}

Neither current-position detection nor the random crosshair differs significantly from the no-cue baseline. By contrast, extrapolating the ball's future position raises success from $12\%$ to $38\%$ relative to current-position detection ($p{=}0.005$), showing that predictive position information is useful for interception. Predicted human gaze achieves the highest observed success rate at $52\%$. Figure~\ref{fig:t6_attention} provides a qualitative view of the corresponding attention patterns.

The timing analysis further distinguishes human gaze from the tested kinematic predictor. The ball reaches the movement-onset gaze location approximately
$541$\,ms after arm movement begins, defining the prediction horizon used by the Kalman cue. Human gaze settles near this location an average of $949$\,ms
before movement onset. When predicting the same ball-arrival event from gaze-dwell onset and movement onset, the Kalman predictor yields mean errors of $420$ and $118$\,px, respectively.

A similar anticipatory pattern appears in the long-horizon Bus Table task (T3). Over more than $1{,}000$ aligned grasp and release events, gaze settles near the next object a median of $1667$\,ms before gripper closure and near the future placement site a median of $933$\,ms before gripper opening; at that time, the manipulated object has not yet reached the placement location. These measurements complement the T6 ablation: human gaze is not merely a marker of the currently visible target, but can also encode upcoming task-relevant locations before they become the current manipulation target.


\section{Limitations}
\label{sec:limitations}

While effective across the six evaluated manipulation tasks, gaze prompting has several limitations:

\begin{itemize}[leftmargin=*,itemsep=2pt,topsep=2pt]

    \item \textbf{Dependence on gaze prediction quality.}
    The inference-time prompt is produced by a learned gaze predictor, so inaccurate predictions can place the crosshair at irrelevant or misleading locations. Although the predictor attains about $10$-pixel held-out localization error on the $224\times224$ policy input, we do not yet characterize systematic or high-confidence prediction errors, nor do we use prediction confidence to suppress unreliable cues.

    \item \textbf{Generalization beyond the evaluated setting.}
    Our experiments use a single robot embodiment, $\pi_0$ backbone, and data-collection setup. Only T1 varies the language instruction within a task, while T2--T6 use a fixed instruction. The current results therefore do not establish generalization to unseen instructions, novel tasks, substantially different visual domains, other robot embodiments, or other VLA backbones.

    \item \textbf{Scope of predictive baselines.}
    The latency-matched current- and future-position baselines are evaluated on T6, where the moving target admits a well-defined kinematic extrapolation. T3 provides complementary timing evidence that gaze can anticipate upcoming object and placement targets, but analogous predictive baselines are less straightforward for long-horizon or multi-object manipulation. A broader characterization of the future task information encoded by gaze remains open.

\end{itemize}


\section{Conclusion}
\label{sec:conclusion}

We introduced \emph{eye-tracker-supervised gaze prompting}, which turns human gaze recorded during VR teleoperation into a visual prompt for VLA fine-tuning and deployment. A lightweight gaze predictor supplies the cue at inference, allowing an unchanged $\pi_0$ policy to use the same input-pixel prompt without an eye tracker or architectural modification. Across six real-world bimanual manipulation tasks, gaze prompting increases mean success from $26.3\%$ to $56.0\%$ under per-task training and retains comparable gains when all six tasks are learned by a single policy.

Controlled analyses further characterize the role and content of the cue. Removing the crosshair only at inference reduces performance, while instruction-swap experiments show stronger instruction-dependent attention under gaze supervision. On the dynamic interception task, predicting the ball's future position is substantially more effective than marking its current position, and predicted human gaze achieves the highest observed success. Timing analyses on both dynamic and long-horizon tasks further show that gaze can indicate upcoming task-relevant locations before they become the current manipulation target. These results support human gaze as a practical source of temporally dense supervision that can be collected alongside demonstrations and injected through the existing visual pathway of a VLA policy.


\clearpage
\acknowledgments{We thank Lingyu Robotics for providing the data, robotic platforms, and technical support. This work was supported by the National Natural Science Foundation of China under Grants 62461160313, 62273196, and 62192752, and by the BNRist project under Grant BNR2024TD03003.}

\bibliography{ref}  

\clearpage
\appendix

\section{Hardware platform}
\label{app:platform}

Demonstrations are collected on TeleAvatar, a commercial bimanual mobile manipulator from LingYu Robotics: two
7-DoF arms, each terminated by a one-DoF gripper, are mounted on a torso column
above a wheeled mobile base.

\paragraph{Sensing.}
The robot carries three RGB cameras---a head-mounted stereo fisheye camera and
one camera on each wrist---all recorded at $30$\,fps. The head camera is captured
side-by-side at $2160\times4320$ ($2160\times2160$ per eye) and the two wrist
cameras at $480\times848$. The head fisheye stream is rectified to an
equirectangular panorama and presented to the operator as a stereo VR180 view
(Appendix~\ref{app:gaze_projection}); the policy consumes the left-cropped square
head view ($2160\times2160$) together with the two wrist views. The robot also
reports proprioceptive joint state, which the policy consumes as its state input.

\paragraph{Teleoperation.}
The operator wears a Pico~4 Enterprise headset and holds its two controllers,
whose poses are retargeted in real time to the two arms and grippers. The
headset's integrated eye tracker simultaneously records the operator's gaze;
Appendix~\ref{app:gaze_projection} describes how this gaze stream is registered
to the head-camera image.


\section{Gaze Data Collection}
\label{app:gaze_projection}

\paragraph{Capture.}
The operator's gaze is recorded by the eye tracker integrated in the Pico~4 Enterprise headset, synchronized with the head-camera stream at $30$\,fps. At each timestep, the tracker emits a 3D gaze ray in the headset frame. Because the policy consumes head-camera pixels rather than headset-frame rays, we register every ray to head-camera image coordinates and Kalman-filter the resulting 2D gaze trace to reduce measurement noise.

\begin{figure}[h]
    \centering
    \includegraphics[width=0.7\linewidth]{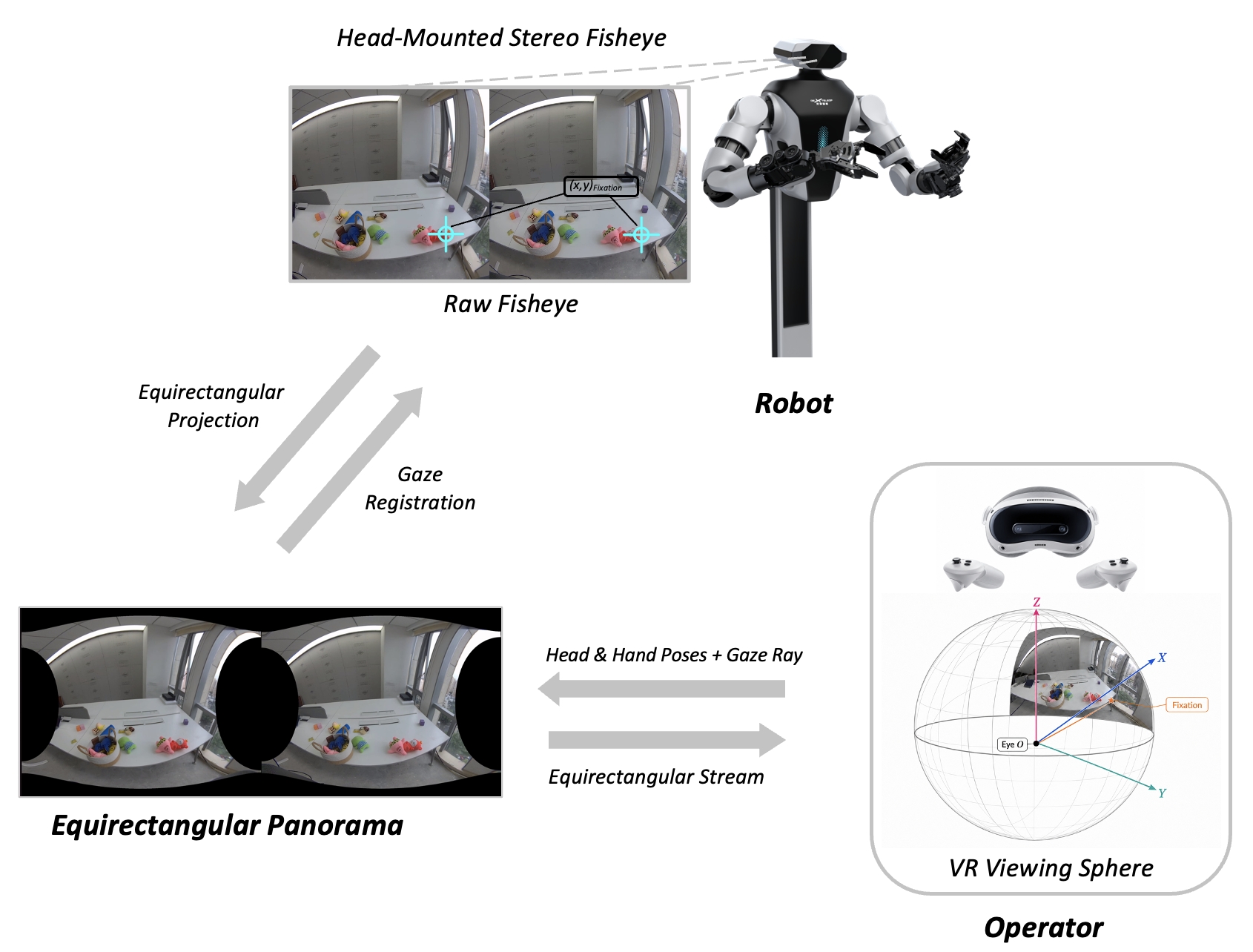}
    \caption{Gaze-to-image registration. The display path maps the head-camera fisheye image to an equirectangular panorama shown on the operator's VR viewing sphere; the gaze-registration path inverts this chain to map the operator's eye-tracker gaze to a 2D location on the head-camera image (\texttt{observation.gaze}).}
    \label{fig:gaze_projection}
\end{figure}

\paragraph{Registration.}
The display path maps a raw fisheye head-camera image to an equirectangular panorama and then onto the operator's VR viewing sphere. Registration inverts this chain. For each frame, we take the gaze direction in the headset frame, intersect this ray with the viewing sphere, convert the intersection to equirectangular coordinates, and map that point back through the calibrated panorama-to-fisheye transformation to a pixel on the original head-camera image. We finally remap this pixel into the left-cropped square view consumed by the predictor and policy, and store the resulting 2D gaze location as \texttt{observation.gaze} (Figure~\ref{fig:gaze_projection}).

Registration runs once, offline, during dataset conversion. The resulting 2D gaze locations supervise the gaze predictor and are rendered as visual prompts during policy fine-tuning. At deployment, the eye tracker is removed: the predictor estimates the gaze location from recent head-camera frames and the language instruction, and the predicted location is rendered in the same way. After dataset conversion, neither predictor training nor policy training/deployment requires headset pose, eye-tracker hardware, camera intrinsics, or viewing-sphere geometry.


\section{Task descriptions}
\label{app:tasks}

We expand on the six \textsc{GazeMani} tasks summarized in
\S\ref{sec:result:setup}. All tasks are collected on the platform described in
Appendix~\ref{app:platform} with 200 demonstrations each. Each task's
instruction is shown in italics.

\paragraph{T1: Stack Bowls.} \emph{(``stack red bowls'' / ``stack yellow bowls'')}
Six same-sized bowls---three red and three yellow---are placed on the table: one
red and one yellow at fixed positions near the robot, and the other two red and
two yellow in a row farther away. The robot stacks the two far bowls of the
instructed color onto the near bowl of that color. A trial is successful when both
far bowls of the instructed color are stacked on the matching near bowl. As the
bowls differ only in color, the task isolates instruction-conditioned instance
selection.

\paragraph{T2: Stack Cubes in Order.} \emph{(``stack cubes in order: yellow first, then purple'')}
A yellow cube and a purple cube are placed on the table together with a marked
target ring. Using both arms, the robot stacks the cubes inside the ring in a
fixed color order, the yellow cube first and the purple cube on top. A trial is
successful when both cubes are stacked in the correct order within the ring. The
per-phase change of target, together with the bimanual handoff, makes this a test
of phase-dependent sequencing.

\paragraph{T3: Bus Table.} \emph{(``bus the table: place dishes, bowls, and cups into the bussing bin, and place trash into the trash bin'')}
The workspace contains tableware---dishes, bowls, and cups---and pieces of trash,
alongside a bussing bin and a trash bin. The robot sorts every item into its
correct bin: tableware into the bussing bin, trash into the trash bin. A trial is
successful when every item reaches its correct bin. With two destinations and many
objects, the task stresses long-horizon target maintenance across an extended
sequence.

\paragraph{T4: Fold Towel.} \emph{(``fold the towel'')}
Starting from a flat towel, the robot folds it once, rotates it, and folds it
twice more to finish. A trial is successful when the towel is fully folded through
this sequence. As the relevant target is a
deforming edge rather than a rigid object, the task tests deformable-state
tracking.

\paragraph{T5: Insert Toilet Paper.} \emph{(``insert the toilet paper roll onto the spindle'')}
A paper roll is placed at a randomized position on the table beside an upright
spindle. The robot picks up the roll and threads it onto the spindle by aligning
the roll's central hole over the tip. A trial is successful when the roll is
seated on the spindle. The small insertion target and the variable starting
position make this a test of precise contact localization.

\paragraph{T6: Catch Rolling Ball.} \emph{(``catch the rolling ping-pong ball and place it into the bin'')}
A ping-pong ball is released on an incline and rolls freely across the workspace.
The robot must intercept the ball before it rolls out of reach and drop it into a
bin. A trial is successful when the ball is caught in time and placed in the bin.
Because the target moves and the grasp is time-critical, the task tests motion
anticipation.


\section{Dataset statistics}
\label{app:dataset_stats}

Table~\ref{tab:dataset_stats} summarizes \textsc{GazeMani}: exactly $200$ demonstrations per task, $1{,}200$ trajectories in total, and $693{,}539$ synchronized frames recorded at $30$\,fps (${\approx}6.4$ hours of teleoperation). Demonstrations are stored in the LeRobot format; every frame carries the full observation--action--gaze tuple described in Appendix~\ref{app:platform}, including the registered 2D gaze location (\texttt{observation.gaze}). Mean episode length spans more than a factor of five across tasks---from the shorter episodes in T5 and T6 (${\sim}7$\,s) to the long-horizon sequences in T3 and T4 ($35$--$40$\,s)---reflecting the differing task structure. Only T1 carries more than one language instruction (``stack red bowls'' / ``stack yellow bowls''), consistent with its role as the instruction-conditioned instance-selection task.

\begin{table}[h]
    \centering
    \small
    \caption{\textsc{GazeMani} dataset statistics, per task and overall.
    Durations assume $30$\,fps; ``Range'' is the shortest--longest episode.
    ``Instr.'' is the number of distinct language instructions.}
    \label{tab:dataset_stats}
    \begin{tabular}{llrrrrr}
        \hline
        Task & Description & Instr. & Episodes & Frames & Mean (s) & Range (s) \\
        \hline
        T1 & Stack Bowls          & 2 & 200 &  77{,}803 & 13.0 &  9.1--19.3 \\
        T2 & Stack Cubes in Order & 1 & 200 &  77{,}307 & 12.9 &  9.8--19.7 \\
        T3 & Bus Table            & 1 & 200 & 211{,}295 & 35.2 & 24.2--70.4 \\
        T4 & Fold Towel           & 1 & 200 & 237{,}817 & 39.6 & 27.3--65.7 \\
        T5 & Insert Toilet Paper  & 1 & 200 &  45{,}288 &  7.5 &  5.7--11.5 \\
        T6 & Catch Rolling Ball   & 1 & 200 &  44{,}029 &  7.3 &  6.0--10.2 \\
        \hline
        \multicolumn{3}{l}{Total} & 1{,}200 & 693{,}539 & \multicolumn{2}{c}{${\approx}385$ min ($6.4$ h)} \\
        \hline
    \end{tabular}
\end{table}


\section{Training and implementation details}
\label{app:training}

Table~\ref{tab:impl} lists the hyperparameters for the gaze predictor and the per-task $\pi_0$ policies. Predictor and policy are trained independently, with no gradients flowing between them (\S\ref{sec:method:training_inference}; predictor architecture in \S\ref{sec:method:predictor}).

\begin{table}[h]
    \centering
    \small
    \caption{Predictor and per-task policy implementation details.}
    \label{tab:impl}
    \begin{tabular}{l p{0.62\linewidth}}
        \hline
        \multicolumn{2}{l}{\textbf{Gaze predictor} (${\sim}2.9$M trainable parameters)} \\
        \hline
        Vision/text encoder & CLIP ViT-B/16 (\texttt{openai/clip-vit-base-patch16}), frozen \\
        Fusion dimension     & $256$ \\
        Temporal window      & $N{=}3$ head frames at $224\times224$ (from the $2160$-px left crop) \\
        Prompt conditioning  & per-frame vision$\to$text cross-attention ($4$ heads) $+$ FFN \\
        Temporal fusion      & $2$-layer transformer encoder ($4$ heads), $N\times196$ tokens \\
        Supervision          & per-frame cross-entropy on Gaussian gaze heatmap (default $\sigma$: Appendix~\ref{app:sigma}) \\
        Optimizer            & AdamW, $\beta{=}(0.9,0.999)$, weight decay $0.01$ \\
        Learning rate        & $1\times10^{-4}$, cosine, $200$ warm-up $\to 1\times10^{-6}$ \\
        Batch / precision    & $64$ / bfloat16 \\
        Sampling             & $25$--$80$ sampled timesteps per episode (scaled with episode length); task-balanced weighted sampler; $80/20$ episode-level train/validation split \\
        Training length      & ${\le}100$ epochs, early stopping (patience $10$) \\
        Inference smoothing  & causal EMA, $\alpha{=}0.2$ \\
        \hline
        \multicolumn{2}{l}{\textbf{$\pi_0$ policy} (full fine-tune, per task)} \\
        \hline
        Initialization       & released $\pi_0$ base~\citep{black2024pi_0}: PaliGemma VLM $+$ 300M action expert \\
        Observations         & head $+$ two wrist views at $224\times224$; crosshair on head only (if any) \\
        Optimizer            & AdamW, $\beta{=}(0.9,0.95)$, weight decay $1\times10^{-10}$, grad-norm clip $1.0$ \\
        Learning rate        & peak $2.5\times10^{-5}$, $1{,}000$ warm-up, cosine decay over $30{,}000$ steps (to $2.5\times10^{-6}$; training stops at $20{,}000$) \\
        Batch / precision    & $64$ / bfloat16 \\
        Weight EMA           & decay $0.99$ \\
        Training length      & $20{,}000$ steps \\
        Action head          & chunk length $30$ \\
        Denoising            & $10$-step flow matching (Euler) \\
        \hline
    \end{tabular}
\end{table}

\paragraph{Unified multi-task policies.}
For the unified evaluation in Figure~\ref{fig:multitask}, each method is trained as a single policy on the pooled demonstrations from all six tasks rather than as six separate policies. We use the same policy hyperparameters as in the per-task experiments, including the optimizer, learning-rate schedule, batch size, and $20{,}000$ training steps. We do not apply task-balanced reweighting when sampling the pooled multi-task data. Evaluation uses $20$ rollouts per task, for $120$ rollouts per method.

\paragraph{Aux-loss baseline.}
Aux-loss uses the Gaze-Regularized VLA objective~\citep{pani2026gaze}, replacing its synthetic gaze with our recorded human gaze while keeping the original KL coefficient. Gaze supervision is used only during training; no gaze cue is provided to the aux-loss policy at inference.

\paragraph{Reproducibility.}
All reported policy cells use one training seed. Per-task success is evaluated over randomized real-robot rollouts as specified in the main text. The \textsc{GazeMani} dataset and evaluation code are available at \url{https://gazemani.github.io/}.


\section{Gaze predictor accuracy and smoothing ablation}
\label{app:sigma}

The predictor is supervised with a Gaussian-target heatmap: ground-truth gaze is rasterized onto the predictor's $224\times224$ output grid and smoothed with a Gaussian of standard deviation $\sigma$ (measured in grid cells; one cell ${\approx}9.6$ px in the $2160$-px head crop). Small $\sigma$ produces a sharp target, whereas large $\sigma$ produces a more spatially diffuse target.

We sweep $\sigma \in \{4,8,16,24\}$ on the joint six-task predictor, each an independent fine-tune from the same initialization on the same demonstrations with all other hyperparameters fixed. Table~\ref{tab:sigma} reports predictor pixel-L2 error in the $2160$-px head-crop coordinates, evaluated with causal EMA smoothing ($\alpha{=}0.2$) and the soft-argmax readout on the held-out validation episodes of each task.

\begin{table}[h]
    \centering
    \small
    \caption{Effect of gaze-heatmap smoothing $\sigma$ on predictor pixel-L2 error (px, lower is better) per task. $\sigma{=}16$ minimizes the mean and is used in all main-paper experiments.}
    \label{tab:sigma}
    \begin{tabular}{lcccccc|c}
        \hline
        $\sigma$ & Catch & Insert & Bowls & Cubes & Bus & Towel & Mean \\
        \hline
        4  & 180.3 & 94.4 & 79.2 & 105.4 & 98.0 & 62.5 & 103.3 \\
        8  & 179.3 & 95.7 & 70.7 & 117.5 & 94.7 & \textbf{54.1} & 102.0 \\
        \textbf{16} & \textbf{165.8} & \textbf{90.6} & \textbf{65.3} & \textbf{105.3} & 92.4 & 61.5 & \textbf{96.8} \\
        24 & 167.2 & 99.2 & 67.1 & 111.2 & \textbf{90.0} & 62.2 & 99.5 \\
        \hline
    \end{tabular}
\end{table}

With the default $\sigma{=}16$, the mean localization error is $96.8$ px in the $2160$-px head crop, corresponding to approximately $10.0$ px after resizing to the policy's $224\times224$ input. This is smaller than the rendered crosshair half-arm of approximately $12$ px at policy resolution (Appendix~\ref{app:crosshair}). The dynamic Catch task is the hardest for the predictor, while the remaining tasks have lower localization error.

The mean error is U-shaped across the tested smoothing values: $\sigma{=}16$ minimizes the average and is best on four of the six tasks (Catch, Insert, Bowls, Cubes); Bus is best at $\sigma{=}24$ by $2.4$ px and Towel at $\sigma{=}8$ by $7.4$ px. Validation cross-entropy is not compared across $\sigma$, because changing the target width changes the entropy of the target distribution itself; we therefore report pixel-L2 localization error.


\section{Visual-prompt rendering details}
\label{app:crosshair}

The gaze prompt is rendered as a fixed crosshair centered at the 2D gaze location on the head-camera image. The prompt is applied only to the head view, since gaze is registered in the head-camera coordinate system; the two wrist-camera views are passed to the policy unchanged.

The same crosshair geometry and appearance are used across all prompted variants, tasks, and instructions: a cross with a small center circle, drawn with a half-arm of $120$ px in the $2160$-px head crop (${\approx}12$ px in the policy's $224\times224$ input), with fixed opacity and a fixed cyan color, RGB $(0,255,255)$.

The color is not conditioned on task identity, instruction, object class, cue source, or predictor confidence. Thus, the prompt supplies spatial information through its location rather than semantic information through task-specific appearance. We use a multi-pixel cross rather than a single-pixel dot so that the cue remains visible after image resizing and visual-token embedding.


\section{Cross-attention extraction from $\pi_0$}
\label{app:attention_extraction}

$\pi_0$ uses the PaliGemma~\citep{beyer2024paligemma} backbone, in which image patch tokens and language tokens are processed by a single self-attention stack; the ``cross-attention'' we visualize is the sub-block of that self-attention whose queries are the instruction tokens and whose keys are the head-camera (\texttt{base\_0\_rgb}) visual tokens. The SigLIP encoder produces a $16\times16$ grid of visual tokens per view, and we use only the head-camera block since the wrist views carry no crosshair. We use the instruction (language) tokens as queries, which directly probe how the instruction grounds to image regions.

We aggregate attention over the intermediate decoder layers $\ell \in \{8,\dots,13\}$ of the $18$-layer Gemma backbone, a band fixed on the vanilla policy and applied identically to all methods. Early layers are close to uniform, and late layers ($\ell \ge 14$) collapse onto attention-sink / register tokens~\citep{xiao2023streamingllm,darcet2024registers} and produce sharp but content-independent maps; the intermediate band is where instruction-conditioned spatial grounding concentrates.

At each frame we run one forward pass on the head-view RGB observation and the tokenized instruction, average the post-softmax weights over query heads and over the real instruction tokens, restrict to the head-camera key columns, average over the layer band, and reshape to $16\times16$.


\section{Position-based cue baselines}
\label{app:cue_baselines}

The cue-source ablation in \S\ref{sec:result:ablation} compares predicted human gaze with position-based cues on T6 (\emph{Catch Rolling Ball}) while keeping the input-pixel rendering route fixed. All prompted variants use the same crosshair geometry and appearance (Appendix~\ref{app:crosshair}).

\paragraph{Current-position cue.}
We use a CPU color-blob detector on the policy's $224\times224$ head-camera input and render the crosshair at the detected ball center in the current frame. The detector runs in $4.06$\,ms per frame, close to the gaze predictor's $3.8$\,ms latency. In a manual check of $200$ pre-grasp frames, the rendered cue lies on the ball in $94.5\%$ of frames (Wilson 95\% CI $[90.4,96.9]$).

\paragraph{Kalman future-position cue.}
We track the detected ball center with a constant-velocity Kalman filter with state
$\mathbf{x}=[x,y,v_x,v_y]^\top$, where position is represented in pixels in the
$2160\times2160$ left-cropped head view and velocity in pixels per second.
Measurement noise is fixed to
$\mathbf{R}=\mathrm{diag}(16,9)$, estimated from detector jitter using a robust
MAD estimate (approximately $4$ and $3$\,px standard deviation along the two image axes).
Process noise follows the standard white-acceleration constant-velocity model with
$\sigma_a=800$; this value is selected on demonstration trajectories to minimize
$541$\,ms extrapolation error.

Measurements whose innovation exceeds $200$ pixels are rejected to suppress occasional
detector jumps to distractors, missing measurements are coasted for at most $10$ frames,
and the actual inter-frame interval is supplied at every update rather than assuming a
fixed frame rate. The future cue is obtained by propagating the filtered state
$0.541$\,s forward and rendering the crosshair at the predicted position. The same
implementation is used to precompute training cues and to generate cues online at
inference. An equivalence check over $30$ episodes gives a maximum discrepancy of
$0.706$\,px between the two pipelines, attributable to serialization rounding.

\paragraph{Random and no-cue controls.}
The random-crosshair control samples the cue location uniformly over the full
$224\times224$ head-camera input, preserving the presence and appearance of the visual
mark while removing task-relevant spatial information. The no-cue control uses the
unmodified head-camera image.


\section{Predictor latency}
\label{app:latency}

The gaze predictor adds one streaming forward pass per policy step. We benchmark wall-clock latency on a single NVIDIA RTX 4090, averaging over $100$ warmed-up iterations. Its streaming forward pass costs $3.8 \pm 0.1$\,ms per frame at steady state, using a rolling cache for the recent frame window and encoding the instruction once.

A single $\pi_0$ policy step takes $70.5 \pm 3.0$\,ms on the same deployment setup. Run serially, gaze prediction, crosshair rendering, and the policy require approximately $75$\,ms per step. In deployment, gaze prediction and rendering run on a background thread off the policy's critical path, so policy execution remains the dominant latency.

For the T6 cue-source comparison, both position-based cues use the CPU color-blob detector described in Appendix~\ref{app:cue_baselines}. The detector takes $4.06$\,ms per frame on the policy's $224\times224$ input, closely matching the gaze predictor's $3.8$\,ms cost. This keeps cue-generation latency comparable when contrasting predicted gaze with current- and future-position cues.


\section{Anticipation timing analysis}
\label{app:anticipation}

We measure gaze lead relative to recorded robot actions in the T6 and T3 demonstrations. Robot events are identified automatically from proprioception; gaze lead is computed from the registered head-camera gaze trace.

\paragraph{T6: Catch Rolling Ball.}
Arm-movement onset, $t_{\mathrm{move}}$, is the first of three consecutive frames in which the maximum absolute right-arm joint velocity exceeds $0.5$; it is detected in all $200$ demonstrations. Let $g_0=g(t_{\mathrm{move}})$ denote the gaze location at movement onset. We trace backward to the beginning of the final continuous dwell within $R=200$\,px of $g_0$ on the $2160\times2160$ head-camera view. The mean observed lead is $949$\,ms ($n=181$); in the remaining $19$ episodes, the dwell is already ongoing at the beginning of the recording and its onset cannot be determined.

The ball reaches the movement-onset gaze location on average $541$\,ms after arm movement begins, setting the $0.541$\,s horizon of the Kalman future-position cue. In a separate timing analysis, we predict the same ball-arrival event from gaze-dwell onset and from movement onset, using the corresponding event-specific horizons. On the same episodes, the mean prediction errors are $420$ and $118$\,px, respectively.

\paragraph{T3: Bus Table.}
Gripper closure and opening are detected automatically from the gripper state. The next-object and placement-location anchors are determined from image differences around each event, independently of gaze. For each anchor, we measure the interval from the onset of the final sustained gaze dwell within $R=150$\,px to the corresponding gripper event. A dwell requires at least five consecutive frames.

The median lead is $1667$\,ms before gripper closure for the next object ($n=1194$) and $933$\,ms before gripper opening for the placement location ($n=1093$). The former interval can include visually guided reaching; at the latter time, the manipulated object has not yet reached the placement location.


\section{Qualitative task rollouts}
\label{app:rollouts}

The following figures show representative rollouts of gaze-prompt $\pi_0$ (ours)
on all six tasks. The cyan crosshair on the head-camera view is the rendered
gaze prompt that the policy receives as input. Frames run left to right (top row
first for the two-row strips). Uncut single-take recordings of all rollouts for
every task and method are available on the project website.

\clearpage
\begin{figure}[H]
    \setlength{\parindent}{0pt}
    \textbf{T1: Stack Bowls} \emph{(``stack red bowls'')}\par\smallskip
    \includegraphics[width=\linewidth]{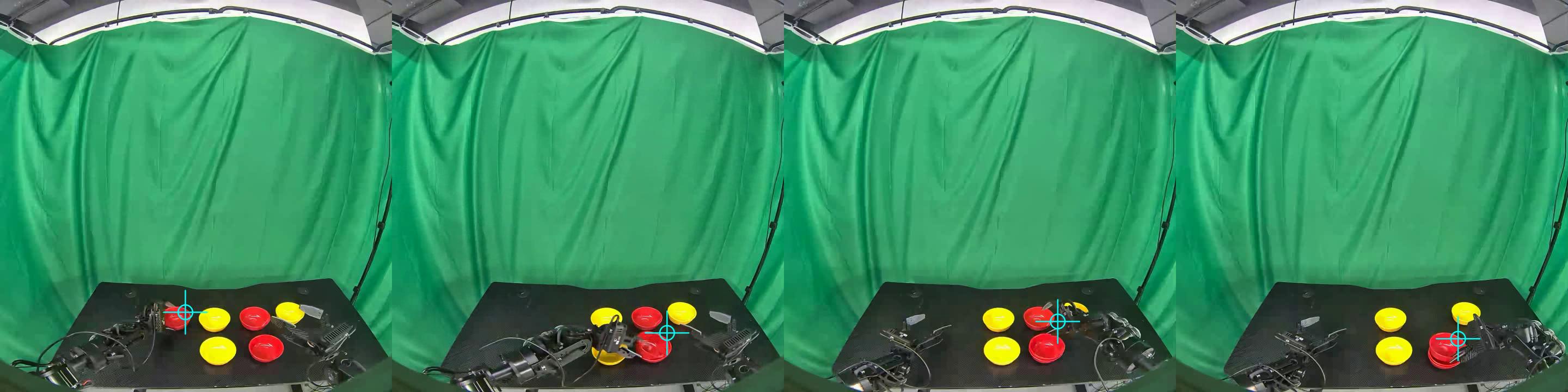}\par\bigskip
    \textbf{T1: Stack Bowls} \emph{(``stack yellow bowls'')}\par\smallskip
    \includegraphics[width=\linewidth]{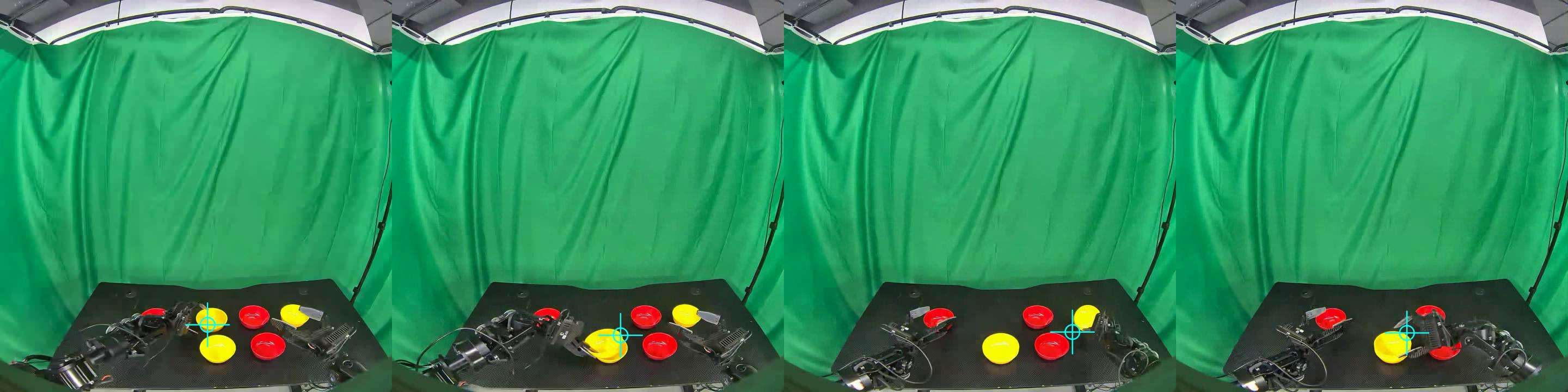}\par\bigskip
    \textbf{T2: Stack Cubes in Order}\par\smallskip
    \includegraphics[width=\linewidth]{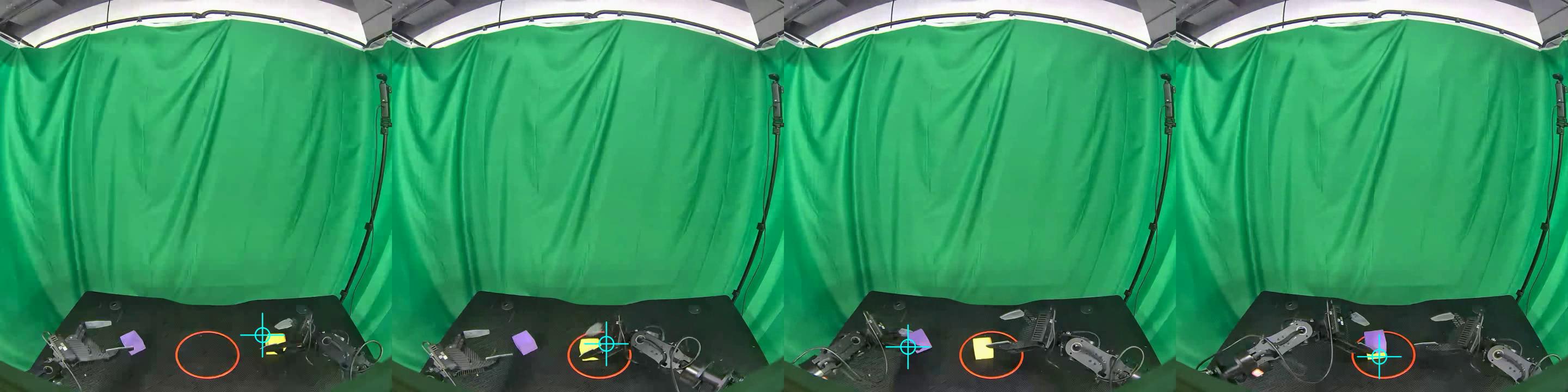}\par\bigskip
    \textbf{T3: Bus Table}\par\smallskip
    \includegraphics[width=\linewidth]{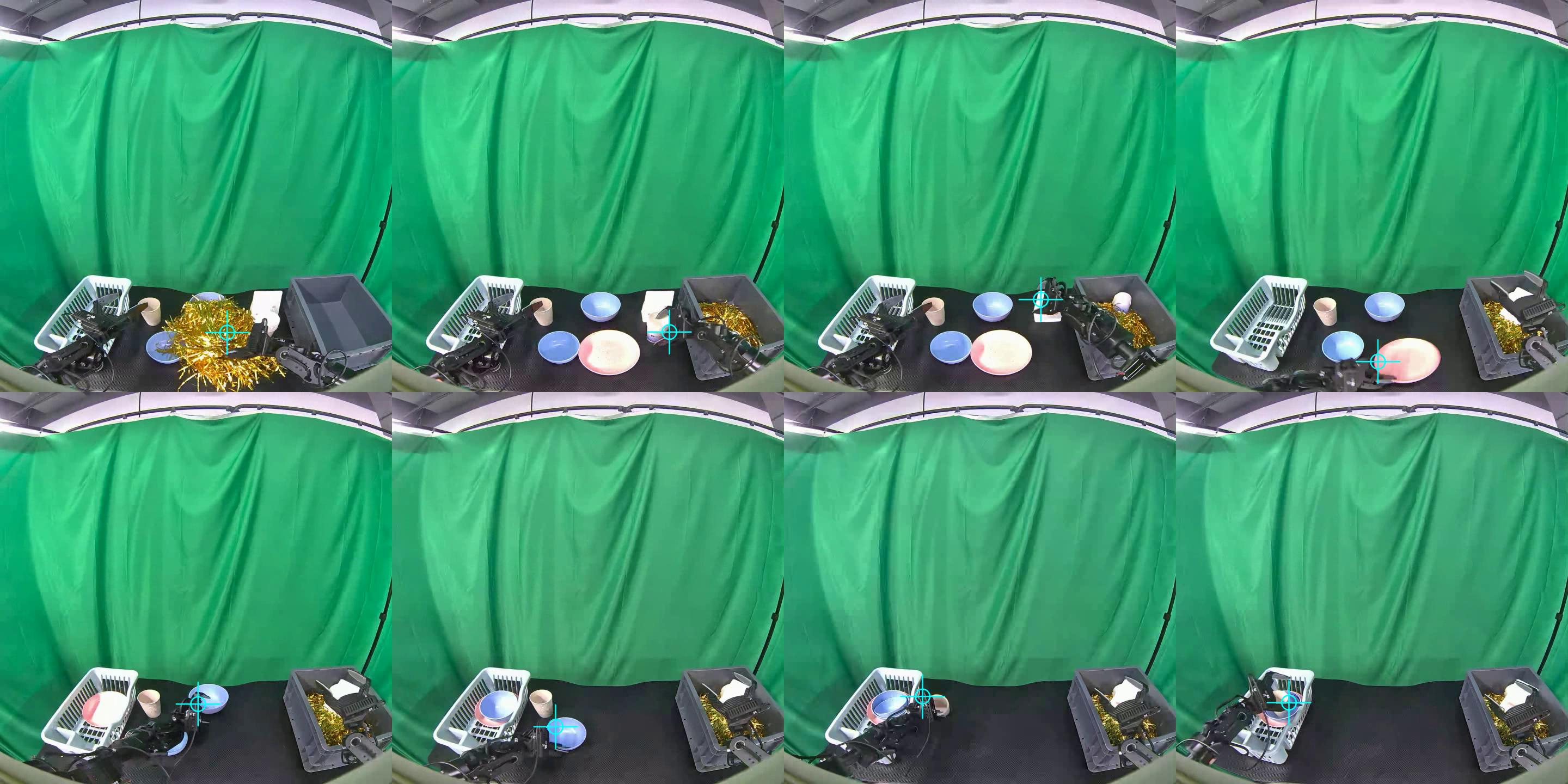}
\end{figure}

\clearpage
\begin{figure}[H]
    \setlength{\parindent}{0pt}
    \textbf{T4: Fold Towel}\par\smallskip
    \includegraphics[width=\linewidth]{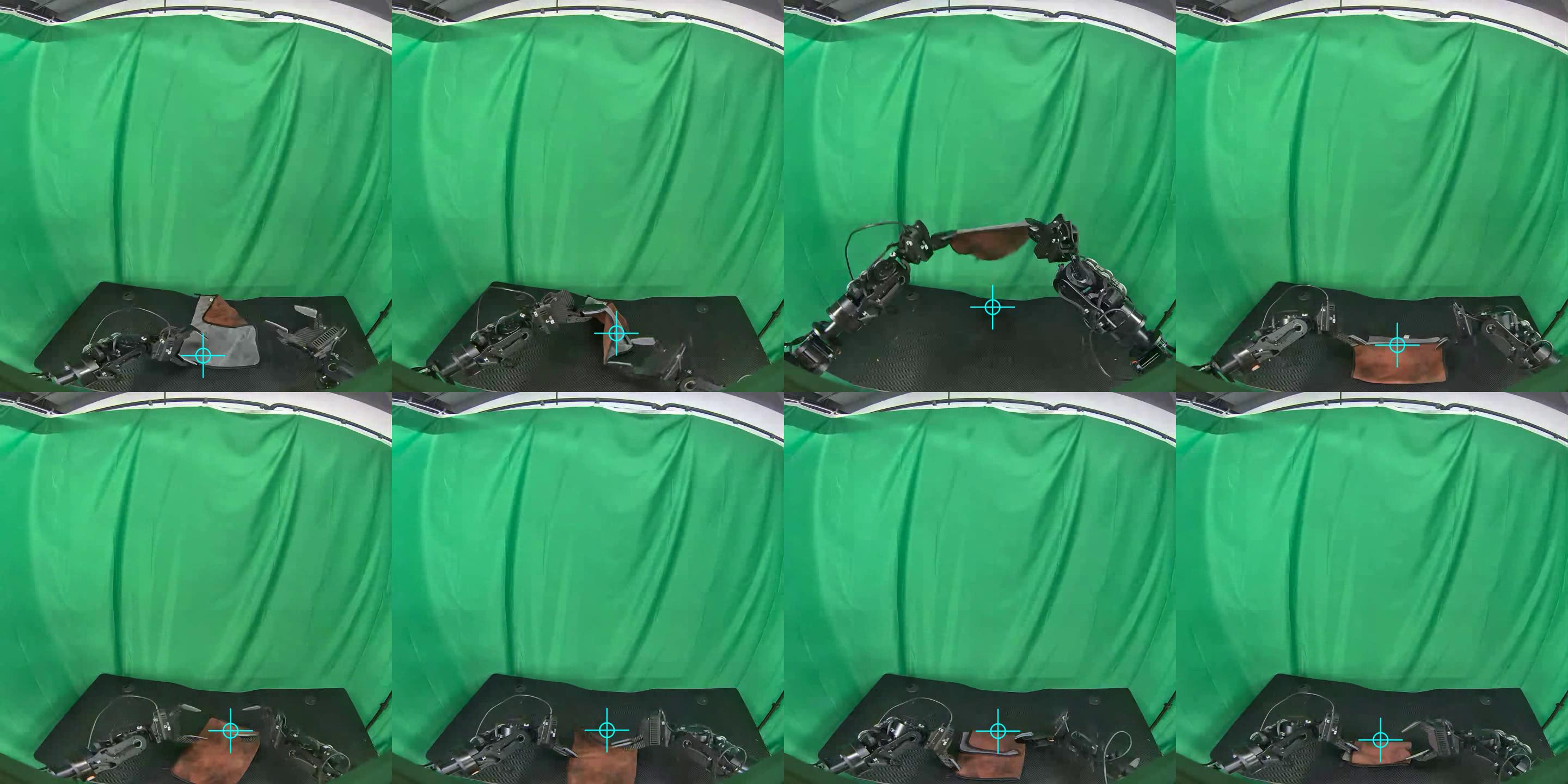}\par\bigskip
    \textbf{T5: Insert Toilet Paper}\par\smallskip
    \includegraphics[width=\linewidth]{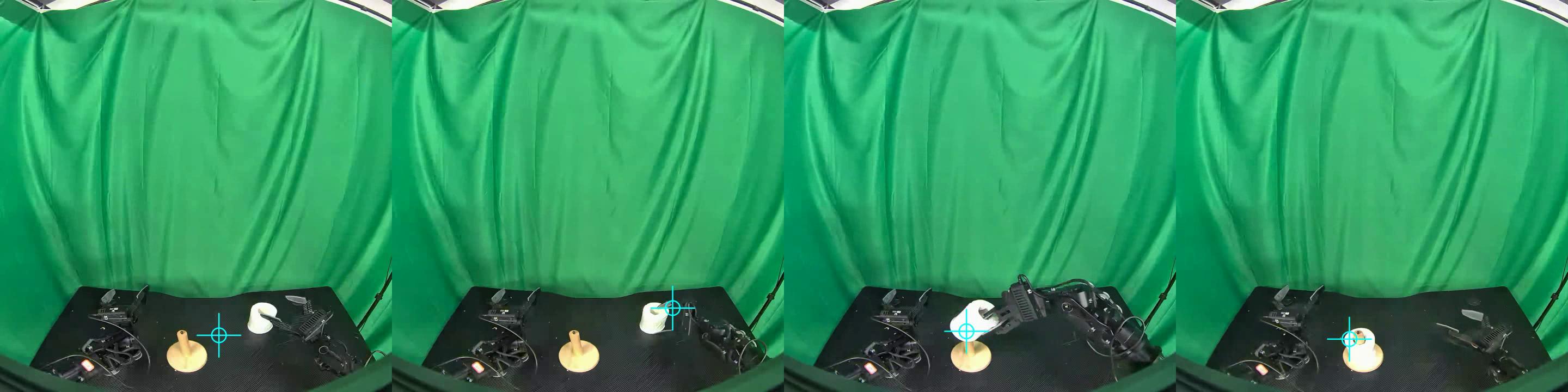}\par\bigskip
    \textbf{T6: Catch Rolling Ball}\par\smallskip
    \includegraphics[width=\linewidth]{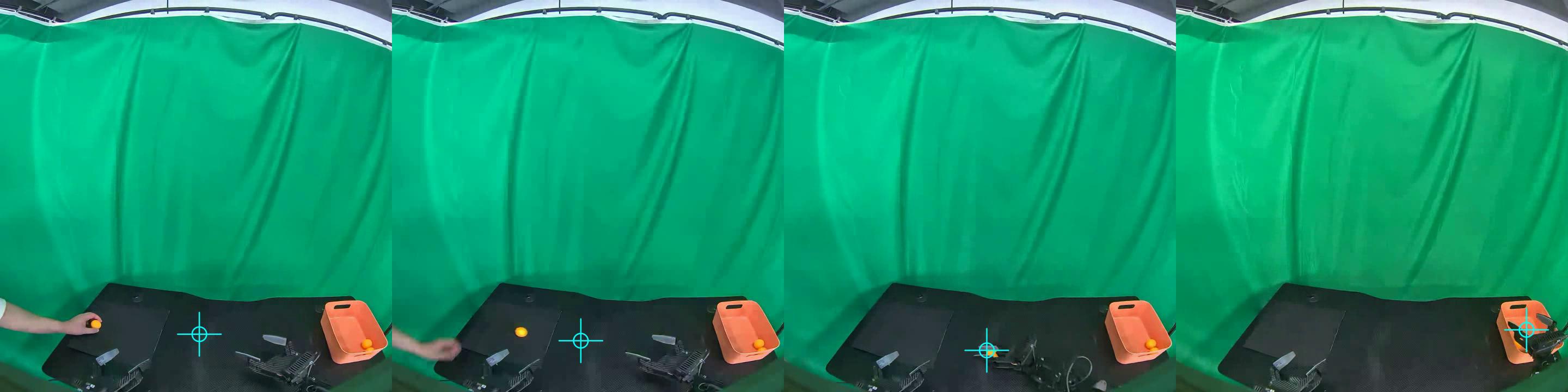}
\end{figure}
\clearpage


\section{Visual robustness}
\label{app:robustness}

We evaluate T1 under two visual changes relative to the data-collection setup.
First, we add previously unseen distractor objects at randomized positions.
Gaze-prompt reaches $68\%$ success, compared with $40\%$ for vanilla $\pi_0$
($n{=}50$ rollouts per method, $p{=}0.009$). Second, we remove the green backdrop
used during data collection. Gaze-prompt reaches $66\%$ success, compared with
$22\%$ for vanilla ($n{=}50$, $p{=}2\times10^{-5}$). For reference, gaze-prompt
reaches $72\%$ and vanilla $38\%$ in the unperturbed T1 evaluation.

These tests show that the performance gain persists under the two evaluated visual
changes. They do not establish generalization to arbitrary lighting, unseen object
instances, or new environments.


\end{document}